\documentclass{article}

\PassOptionsToPackage{numbers, compress}{natbib}

 \usepackage[preprint]{neurips_2026}

\usepackage[utf8]{inputenc} %
\usepackage[T1]{fontenc}    %
\usepackage{hyperref}       %
\usepackage{url}            %
\usepackage{booktabs}       %
\usepackage{amsfonts}       %
\usepackage{nicefrac}       %
\usepackage{microtype}      %
\usepackage{xcolor}         %
\usepackage{graphicx}
\usepackage{amsmath}

\title{SimpleProc: Fully Procedural Synthetic Data from Simple Rules for Multi-View Stereo}

\author{Zeyu Ma \qquad
Alexander Raistrick \qquad
Jia Deng \\
Princeton University
}

\begin{document}

\maketitle

\begin{abstract}
Generating procedural synthetic data for multi-view stereo (MVS) usually requires writing complex rules to match the realism of curated datasets. We demonstrate that we can generate effective training data using SimpleProc: a new, fully procedural generator driven by a very small set of rules based on Non-Uniform Rational Basis Splines (NURBS), as well as simple displacement and texture patterns.
At a modest scale of 8,000 images, our approach achieves superior results compared to manually curated images at the same scale sourced from games and real-world objects. When scaled to 352,000 images, our approach achieves similar, and in some benchmarks, even better results than the current state-of-the-art trained on over 692,000 manually curated images. The source code and the data are available at \url{https://github.com/princeton-vl/SimpleProc}.

\end{abstract}

\begin{figure}[h!]
  \centering
  \includegraphics[width=\linewidth]{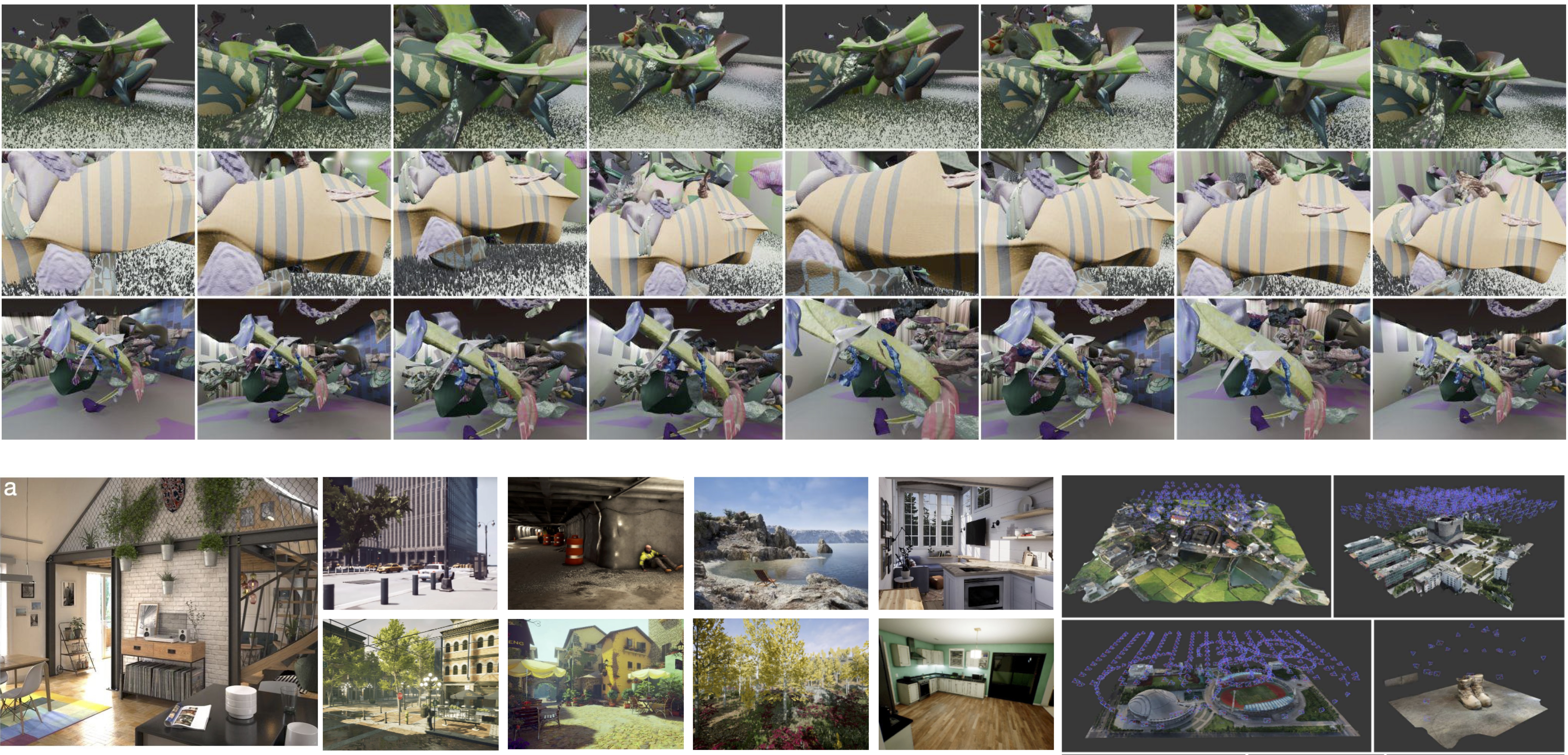}
  \caption{Fully procedural synthetic data from simple rules (top) is as effective as curated data from artists or 3D scans (bottom) for training multi-view stereo models.}
  \label{fig:teaser}
\end{figure}

\section{Introduction}

Synthetic data from computer graphics is widely used in 3D vision. Optical flow models such as WAFT~\cite{wang2025waft} and FlowFormer++~\cite{shi2023flowformer} rely on FlyingChairs~\cite{dosovitskiy2015flownet}, FlyingThings3D~\cite{mayer2016large}, Sintel~\cite{sintel}, and Spring~\cite{spring}. Multi-view stereo models and monocular depth models such as MVSAnywhere \cite{izquierdo2025mvsanywhere} and Depth Anything V2 \cite{yang2024depthv2} often rely on BlendedMVS \cite{yao2020blendedmvs}, HyperSim \cite{roberts2021hypersim}, and other synthetic datasets. Such synthetic data provides accurate geometric ground truth to supervise training.

Generating synthetic data using computer graphics requires a large dataset of 3D assets. A common approach is to use assets created by artists~\cite{roberts2021hypersim,wang2020tartanair} or from video games~\cite{hu2021sailvos}. Another approach is to scan and reconstruct real-world objects~\cite{yao2020blendedmvs}. However, these methods are labor-intensive, as they require significant human effort for each additional asset.

An alternative approach is procedural generation, which creates data entirely through mathematical algorithms and rules. Procedural dataset generation has several advantages: it offers an infinite variety of randomized outputs, it allows full control of dataset properties, and it can generate large, diverse datasets at minimal cost.

A fundamental challenge in procedural generation is how to design the procedural rules. Given the vast design space of procedural rules, we ask: What kinds of rules are needed? Do we need rules to cover many different categories of objects? What rules are effective for different tasks? These questions remain understudied.

In this paper, we explore the design space of procedural rules for multi-view stereo (MVS), a fundamental task in 3D vision. MVS aims to reconstruct a 3D scene using a collection of images taken from multiple camera views. It has many downstream applications, including autonomous driving, robotics, and augmented reality (AR). MVS is one of the most common configurations for dense 3D reconstruction whenever as few as two camera views are available.

We examine the hypothesis: Does there exist a small set of rules that can generate effective training data? The question is intriguing because the typical training data for multi-view stereo covers diverse realistic scenes. To achieve realism in procedural generation, one would need many rules to cover a wide range of object categories in the real world. On the other hand, there is reason to believe that realism is unnecessary, as a general MVS system should reconstruct \emph{arbitrary} shapes with \emph{arbitrary} materials, not just shapes of particular object categories like cars or laptops.

We are not the first to pose this hypothesis. MegaSynth~\cite{jiang2025megasynth} provides large-scale synthetic data with procedural shapes, and Shape Evolution~\cite{yang2018shape} jointly generates synthetic shapes while training a deep network. However, they address different questions. First, their tasks differ: MegaSynth focuses on novel view synthesis (and performs significantly worse than our approach when applied to MVS; see Sec.~\ref{sec:same_budget}); and Shape Evolution focuses on the shape-from-shading task. Second, MegaSynth is not fully procedural: it uses a library of non-procedural image textures for materials.

In this paper, we demonstrate that we can generate effective training data for multi-view stereo using a small set of simple rules. Specifically, we introduce SimpleProc, a procedural generator that constructs shapes as Non-Uniform Rational Basis Splines (NURBS) surfaces through a lofting process, derives textures and materials from simple patterns such as Perlin noise, and arranges these shapes at various scales within the scene (see the top row of Fig.~\ref{fig:teaser}).

We evaluate our generated data by training MVSAnywhere and benchmarking on the Robust Multi-View Depth (RMVD) benchmark~\cite{schroppel2022benchmark}.
Under a fixed data budget of 8,000 images, our data achieves superior results compared to both MegaSynth and the curated 8-dataset baseline.
By scaling the data budget to 352,000 images, we achieved similar, and in some benchmarks, even better results compared to the current state-of-the-art trained on over 692,000 manually curated images.

Our contributions are twofold: (1) we show that state-of-the-art MVS performance can be reached from data based on a small set of simple rules; (2) we provide a procedural generator and a large-scale dataset for MVS.

\section{Related Work}
\subsection{Multi-View Stereo}
Multi-View Stereo (MVS) has many important applications including autonomous driving, robotics, and augmented reality (AR), and it is one of the most common settings requiring as few as two camera views. 
Many approaches focus on optimizing performance in individual benchmarks, e.g., works in the vein of MVSNet~\cite{yao2018mvsnet} focus on DTU~\cite{aanaes2016large} and Tanks \& Temples~\cite{knapitsch2017tanks}, while those in the vein of PatchMatchNet~\cite{wang2021patchmatchnet} focus on the performance on ETH3D~\cite{schops2017multi}.
Subsequently, the Robust Multi-view Depth (RMVD) benchmark~\cite{schroppel2022benchmark} was established, which consists of five diverse datasets: KITTI~\cite{Geiger2012CVPR}, ScanNet~\cite{dai2017scannet}, ETH3D, DTU, and Tanks \& Temples. To achieve robustness on RMVD, MVSAnywhere~\cite{izquierdo2025mvsanywhere} relies on large mixtures of curated training data.  Rather than proposing architectural modifications, we choose MVSAnywhere as our baseline model and focus purely on the data side, investigating whether minimalist procedural data can replace manual curation.

\subsection{Synthetic Data for MVS}

\label{sec:rel-synth}

Yao \textit{et al.} introduced BlendedMVS~\cite{yao2020blendedmvs},  which has become a standard training recipe for MVS models. BlendedMVS is not purely synthetic as it captures images of real object along manual designed trajectories. Training on it alone is not sufficient for superior performance in RMVD (as proved in the MVSAnywhere paper~\cite{izquierdo2025mvsanywhere}). To further scale up the amount of training data, MVSAnywhere uses 8 datasets including synthetic data from games: Hypersim~\cite{roberts2021hypersim}, TartanAir~\cite{wang2020tartanair}, BlendedMVS~\cite{yao2020blendedmvs}, MatrixCity~\cite{li2023matrixcity}, VKITTI2~\cite{cabon2020virtual}, Dynamic Replica~\cite{karaev2023dynamicstereo}, MVSSynth~\cite{huang2018deepmvs}, and SAIL-VOS 3D~\cite{hu2021sailvos}. It takes substantial effort to curate such datasets. In contrast, we scale our data in a different direction: the procedural methods.

\subsection{Procedural Data}
Procedural data offers the advantage of generating infinite variation through a compact set of rules. For instance, Infinigen~\cite{raistrick2023infinigen} and its indoor variant~\cite{raistrick2024infinigen} focus on photorealism but are computationally expensive. 

Other frameworks, such as Kubric~\cite{greff2022kubric}, provide procedural pipelines for generating 3D scenes but rely on existing asset libraries. Similarly, MegaSynth~\cite{jiang2025megasynth} provides procedural scenes but uses non-procedural textures.

In addition, MegaSynth focuses on novel view synthesis tasks; it does not ablate the impact of individual procedural rules; and it serves as a complementary part of the training data to the real data, rather than a standalone solution.

In contrast, we aim to demonstrate the effectiveness of minimalist procedural data as the only data source. We design our generator with minimalist principles and detailed ablation experiments.

\section{Task and Base Model}
We focus on the task of multi-view stereo. Given a set of neighboring images $\{I_i \mid i = 0,1, \dots, N\}$ with known camera information, the model outputs a depth map for one of the views called the reference view ($I_0$). Usually, $N$ is smaller than 10, but large-scale scene reconstruction is achieved by fusing many depth maps. RMVD~\cite{schroppel2022benchmark} is a benchmark evaluating the accuracy of the models across five diverse real-world datasets: 
KITTI \cite{Geiger2012CVPR}, ScanNet \cite{dai2017scannet}, ETH3D \cite{schops2017multi}, DTU \cite{aanaes2016large}, and Tanks \& Temples~\cite{knapitsch2017tanks}.

We conduct our experiments using the current state-of-the-art model on this benchmark, MVSAnywhere~\cite{izquierdo2025mvsanywhere} (DoubleTake License). Testing with MVSAnywhere shows that state-of-the-art MVS performance is reachable with simple, fully procedural data. However, our data generator is by no means specialized to MVSAnywhere.

Meanwhile, MVSAnywhere is also a representative case study as many contemporary MVS models share similar frameworks, such as MVSNet~\cite{yao2018mvsnet}, CasMVSNet~\cite{gu2020cascade}, CERMVS~\cite{ma2022multiview}, and MVSFormer++~\cite{cao2024mvsformer++}:

\begin{itemize}
    \item Feature Extraction: Deep features are extracted from all the images using a shared encoder.
    \item Cost-Volume Construction (Global/Local): Depth hypotheses ($D$ in total) are sampled across either the full $d_{\text{min}}$ to $d_{\text{max}}$ range or a local $d_{\text{current}} \pm \Delta d$ range for each pixel in $H \times W$ space. Feature maps are warped according to the depth, and matching scores are computed to form a cost volume of shape $D \times H \times W$.
    \item Depth Prediction: Either 3D convolutional regularization or a GRU unit is used to predict the depth probability distribution or an iterative update to the current depth estimate.
\end{itemize}

In their paper, MVSAnywhere is trained on the curated 8-dataset mixture (692,000 images) listed in Sec.~\ref{sec:rel-synth}.
We aim to replace these datasets with procedural data generated using the pipeline described below.

\begin{figure}[h!]
  \centering
  \includegraphics[width=\linewidth]{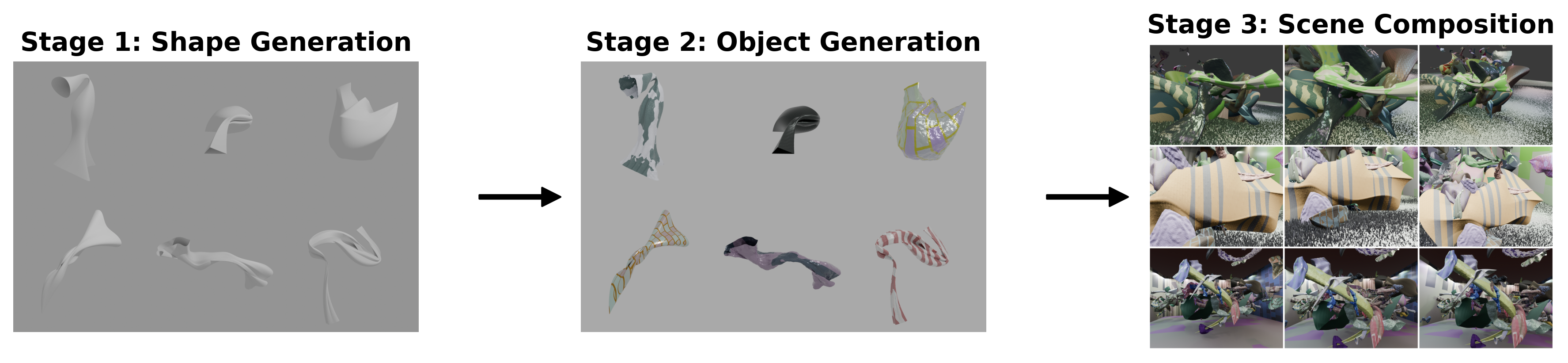}
  \caption{Procedural Data Generation Pipeline. Stage 1 generates NURBS surfaces from the lofting operation. Stage 2 applies displacements, textures, and material properties. Stage 3 arranges cameras, objects, lighting, and optional room boxes.}
  \label{fig:pip}
\end{figure}
\section{Data Generation Pipeline}
\textbf{Overview.} Our data generation pipeline (Fig.~\ref{fig:pip}) is based on Blender \cite{blender} and utility functions from Infinigen~\cite{raistrick2023infinigen}. It consists of multiple stages: Stage 1 - shape generation; Stage 2 - materials and mesh displacement; and Stage 3 - camera placement, object arrangement, lighting, and rendering.

\begin{figure}[b!]
  \centering
  \includegraphics[width=0.48\linewidth]{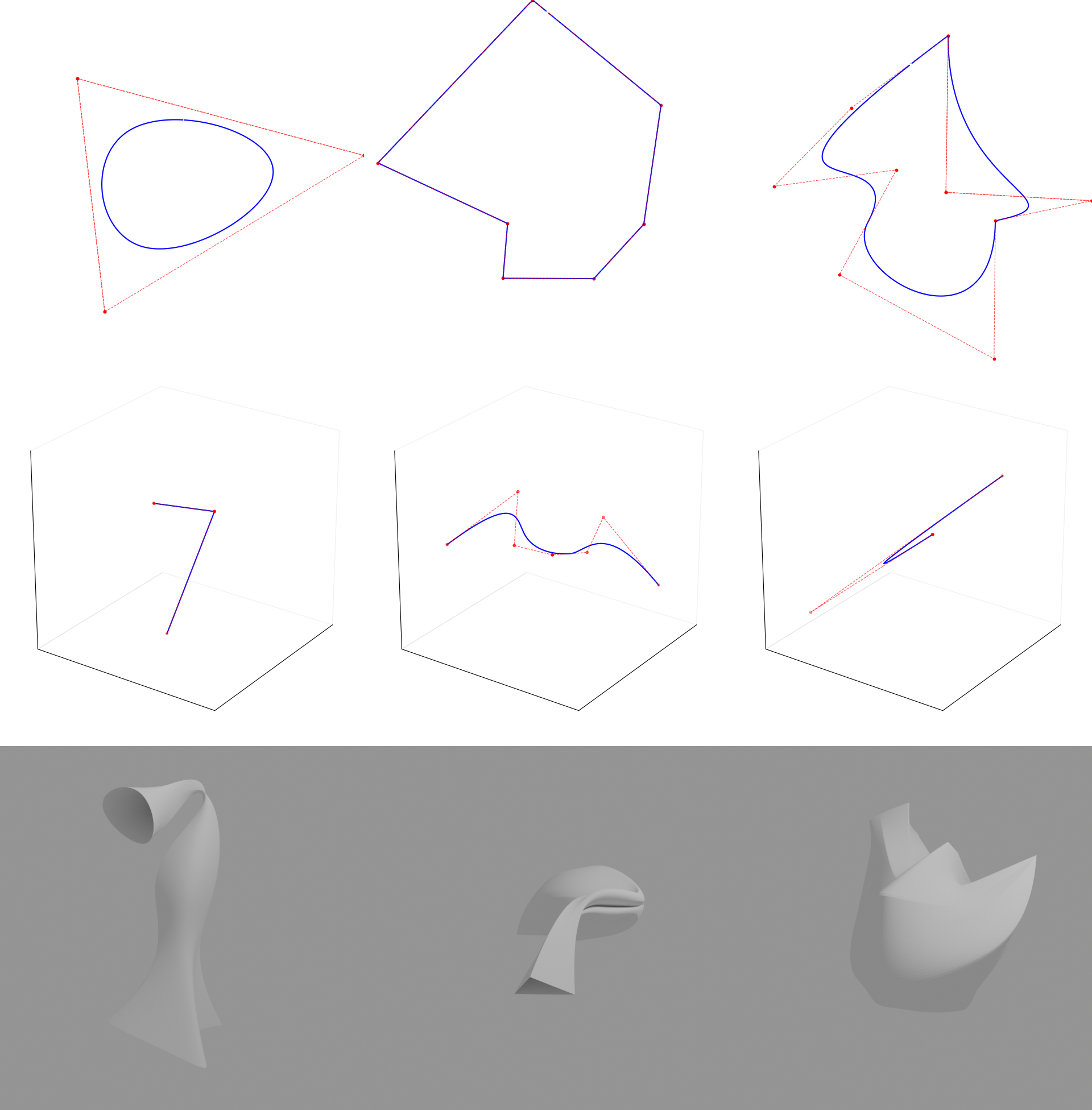}
  \includegraphics[width=0.48\linewidth]{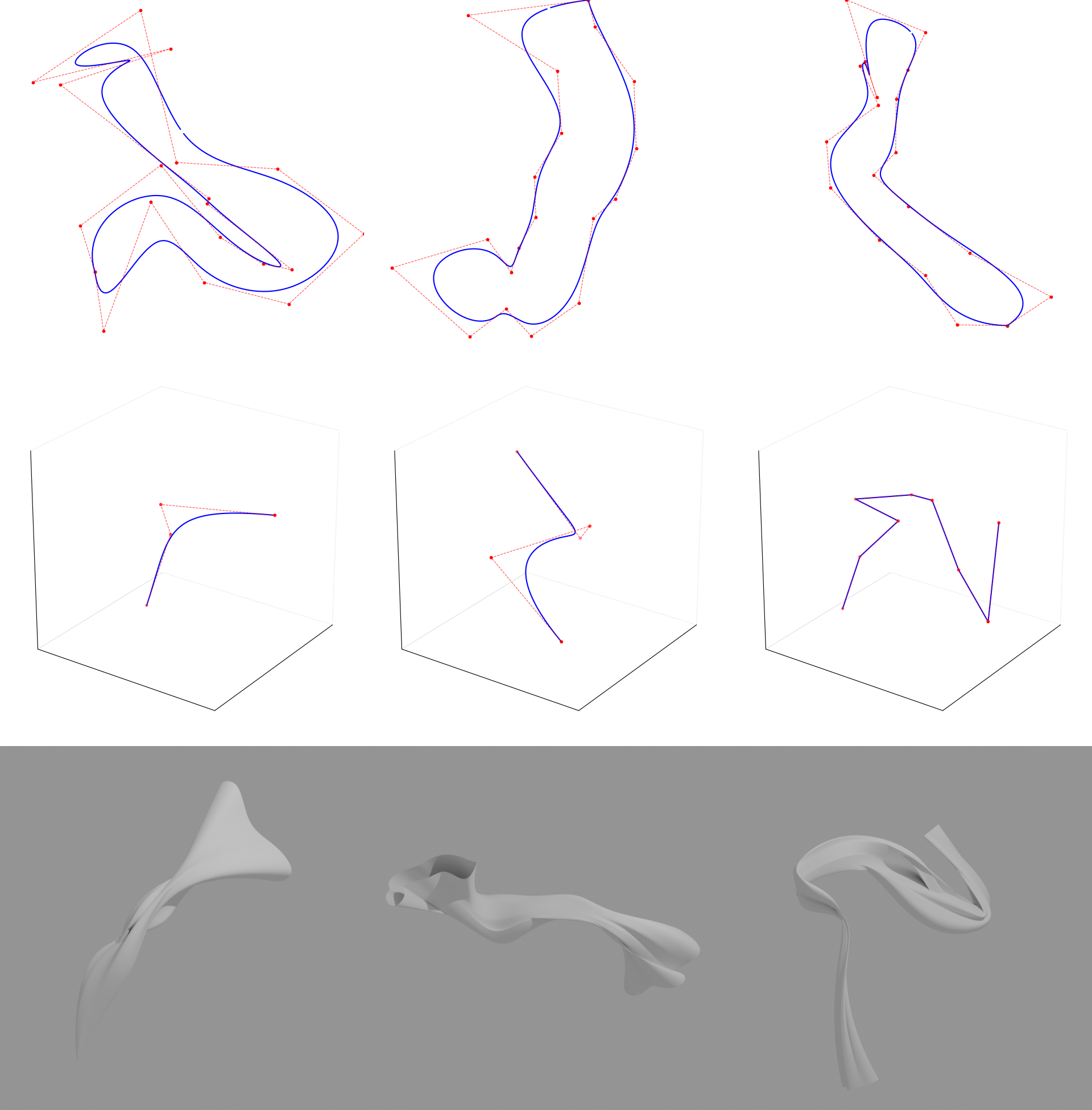}
  \caption{Shape generation pipeline. Top: Profile curves (Starfish and Reptile styles) with control points in red (only one of the profiles is shown for each shape). Middle: Stem curves from 3D random walks. Bottom: Resulting lofted NURBS surfaces showing diverse smooth and sharp features.}
  \label{fig:shapes}
\end{figure}

\subsection{Shapes}
All of our shapes are generated by \textbf{lofting} a \textbf{profile} curve along a \textbf{stem} curve. The profile curve is a closed curve, and the stem curve is an open curve. The lofting process creates a NURBS surface by sampling different profile curves as cross-sections uniformly along the stem curve. A random scaling factor is also applied to the profiles, and the profiles are interpolated smoothly. 

Both the profile and the stem are NURBS curves with a degree between 1 and 3, defined by a set of control points and a knot vector. The stem curve has control points generated from a 3D random walk. The profile curve follows one of two styles:
\begin{itemize}
    \item Starfish style: We generate a set of control points uniformly in a circle, and then perturb each point with independent Gaussian noise in both radial and tangential directions. (Fig.~\ref{fig:shapes}, top-left).
    \item Reptile style: We generate a sequence of points from a 2D random walk and turn it into an open NURBS curve. Then we convert the open curve into a closed profile by offsetting it with a constant radius and fitting a closed NURBS curve to it. (Fig.~\ref{fig:shapes}, top-right).
\end{itemize}

The bottom of Fig.~\ref{fig:shapes}  shows several examples of the generated shapes through this lofting operation. Both smooth and sharp features are present in the shapes.

\subsection{Displacement}
\label{displacement}
We augment the base NURBS shapes with \textit{micro-geometry}, meaning small-magnitude displacement of their vertices. This uses a combination of the two mechanisms below:
\begin{itemize}
    \item Geometry Nodes: for each shape, we use Blender's brick texture, wave texture, or noise texture with random scale and magnitude along the normal direction. To alleviate low-poly artifacts, we use a relatively high resolution in the previous shape generation step followed by the ``subdivide'' operator.
    \item Displacement Socket in Shaders: sometimes, even the "subdivide" operator is too expensive to give sufficient geometry detail. Therefore, we use one of the three noise types in the displacement socket of the object's shader as well. This technique efficiently creates detail at very small scales.
\end{itemize}

The top row of Fig.~\ref{fig:disp_text} shows several example displacements applied to the shapes in Fig.~\ref{fig:shapes}.

\subsection{Texture and Material Properties}
\label{sec:tex}
\begin{figure}[h!]
  \centering
  \includegraphics[width=.48\linewidth]{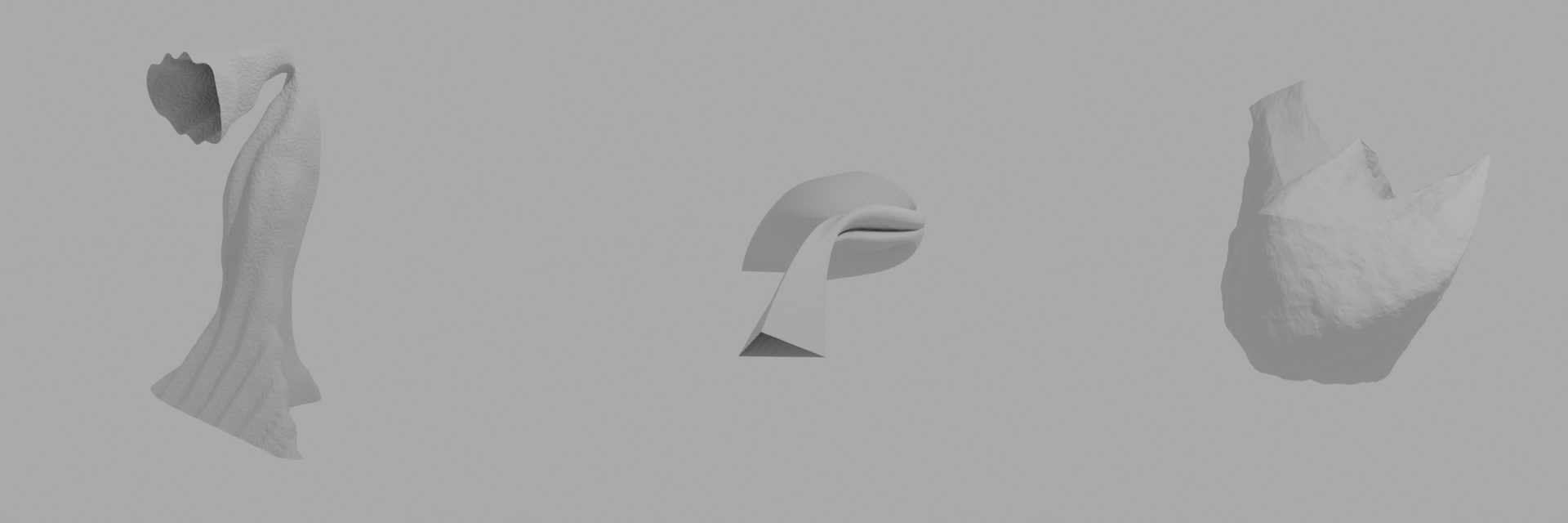}
  \includegraphics[width=.48\linewidth]{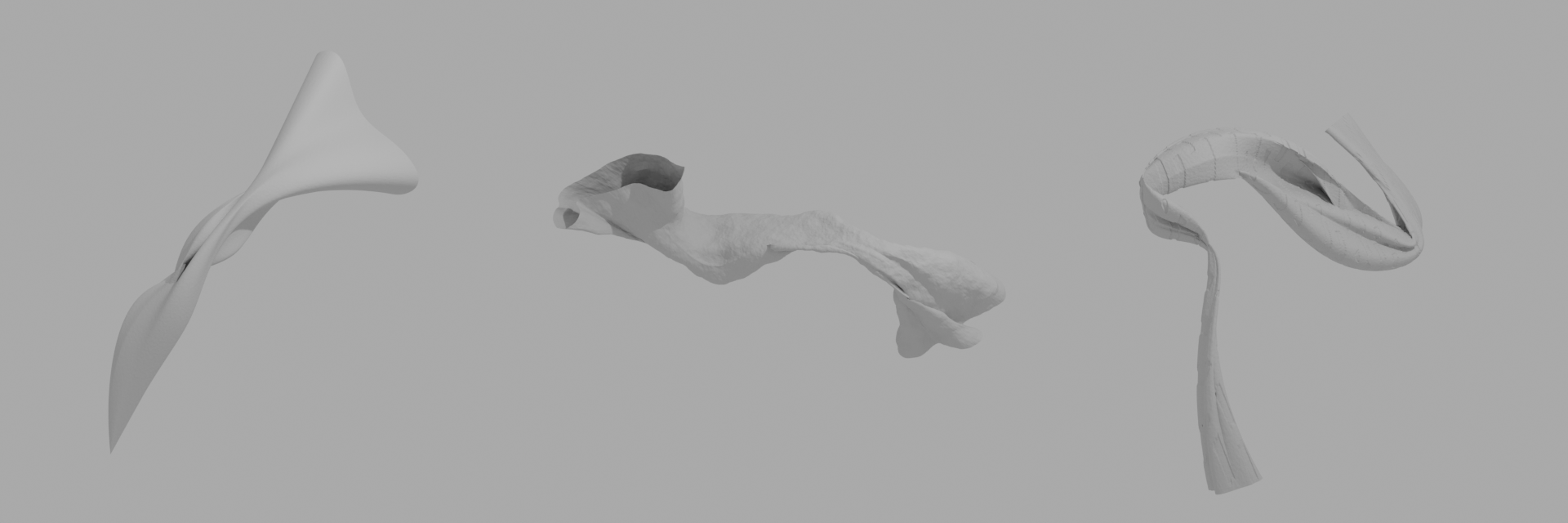}
    \includegraphics[width=.48\linewidth]{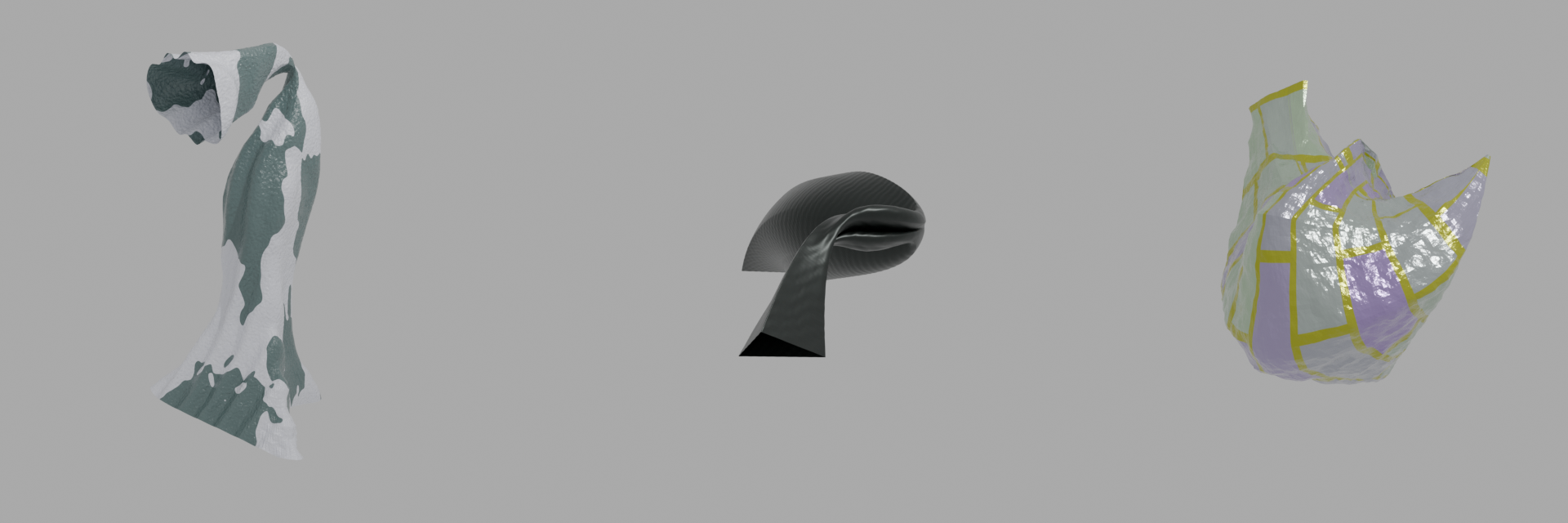}
  \includegraphics[width=.48\linewidth]{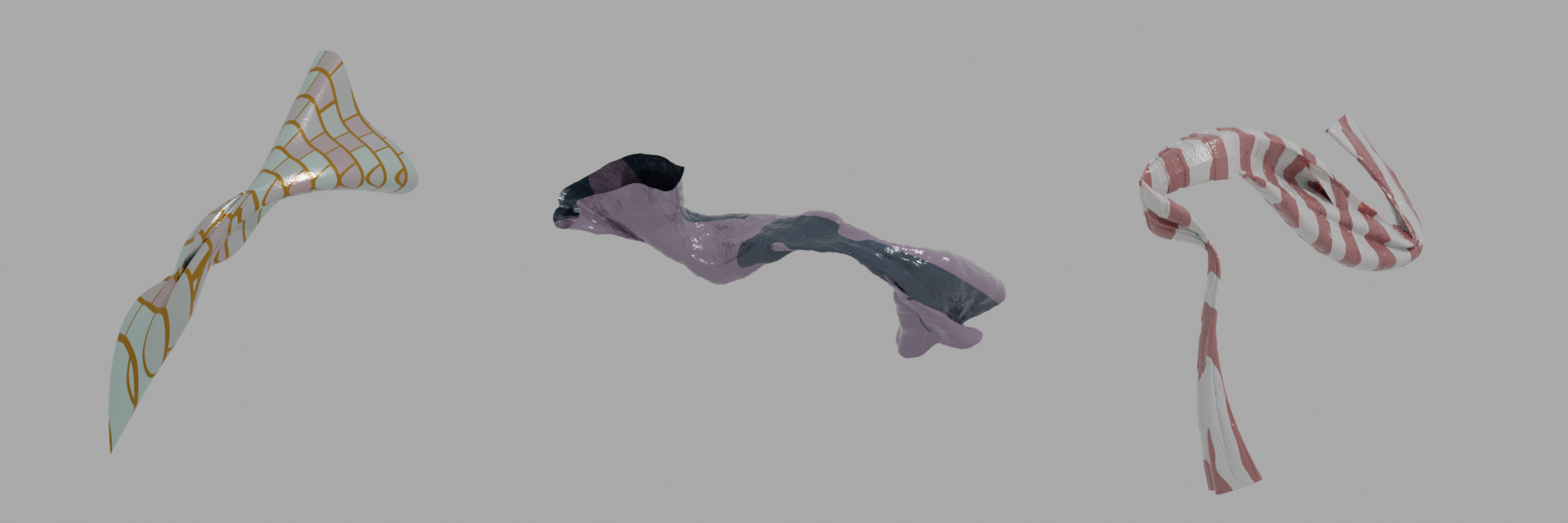}
      \includegraphics[width=.48\linewidth]{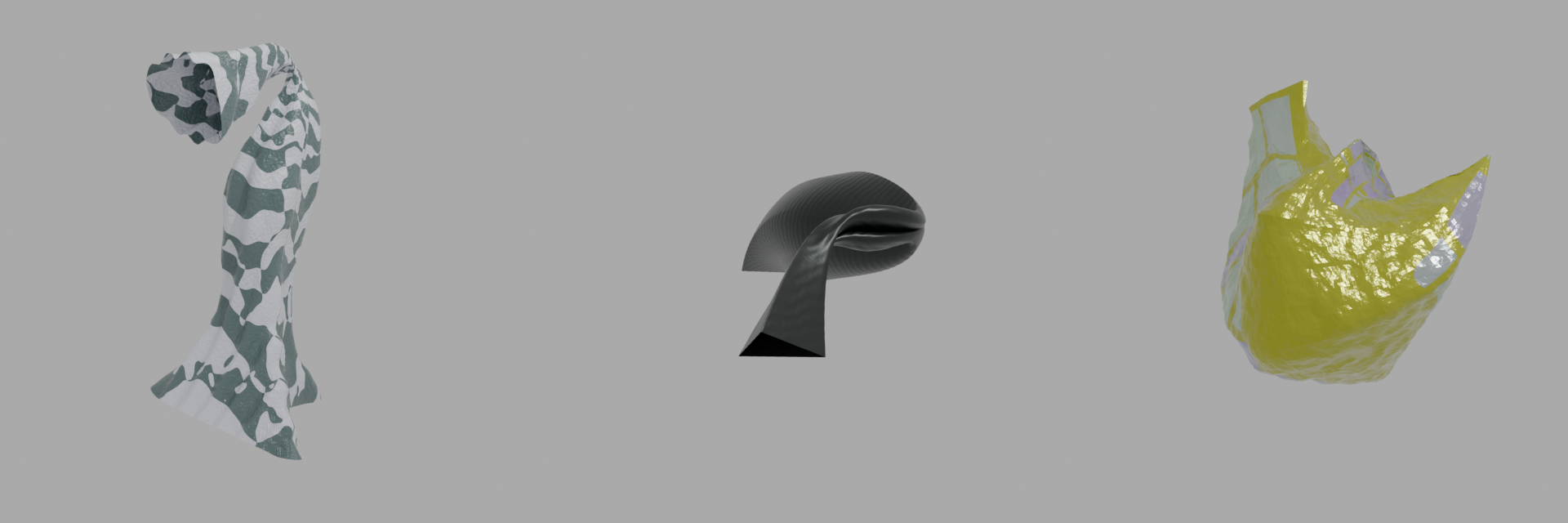}
  \includegraphics[width=.48\linewidth]{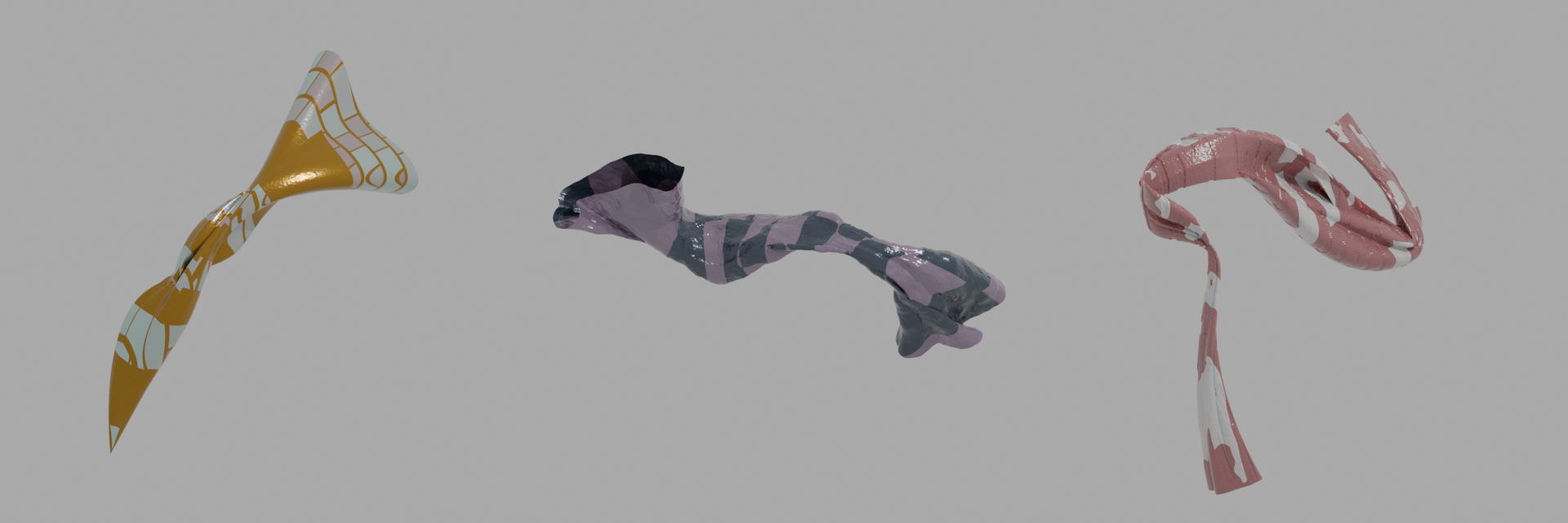}
  \caption{Shapes with Displacement and Textures. The top row shows displacements applied to base shapes. The middle row shows textures from brick, wave, or Perlin noise pattern. The bottom row shows complex textures created by boolean operations.}
  \label{fig:disp_text}
\end{figure}

Similar to the displacement, we use brick, wave, or Perlin noise patterns for the texture. For the wave and noise patterns, we convert their continuous values into two discrete regions using a threshold, and assign a distinct color to each region. Such textures are shown in the middle row of Fig.~\ref{fig:disp_text}. We use Boolean operations to combine two such textures for each object (Fig.~\ref{fig:disp_text}, bottom row); for example, the gray color in the leftmost shape is a boolean XOR of a Perlin noise pattern and a wave pattern.

The color values in the textures are uniformly sampled in HSV space. Note that the rendered images have different HSV distribution depending on other factors, including lighting, rendering engines, and post-processing.

Other material properties are randomly sampled from either a default value or a specified range. For example, roughness is set to 0.2 with 50\% probability and uniformly sampled from $[0.2, 1.0]$ otherwise; metallic is set to 0 with 50\% probability and sampled from $[0, 0.8]$ otherwise.

\subsection{Camera Placement}

To make the objects more efficiently used, we place cameras before adding objects into the scene. The position of the camera is sampled by randomly selecting the azimuth angle, elevation angle, and radius within predefined ranges in spherical coordinates (an elevation $\phi \in [-5^{\circ}, 30^{\circ}]$ and an azimuth $\theta$ spanning $45^{\circ}$). Each camera is oriented to look at the scene center, with the z-axis defined as the up direction, and a small perturbation is applied to the final rotation. We place a total of eight cameras in each scene to maximize the diversity of the dataset. The horizontal field-of-view of each camera is randomly sampled within the range $[31^{\circ}, 97^{\circ}]$, and the aspect ratio is set to $3:4$ (height:width).

\begin{figure}[b!]
  \centering
  \includegraphics[width=\linewidth]{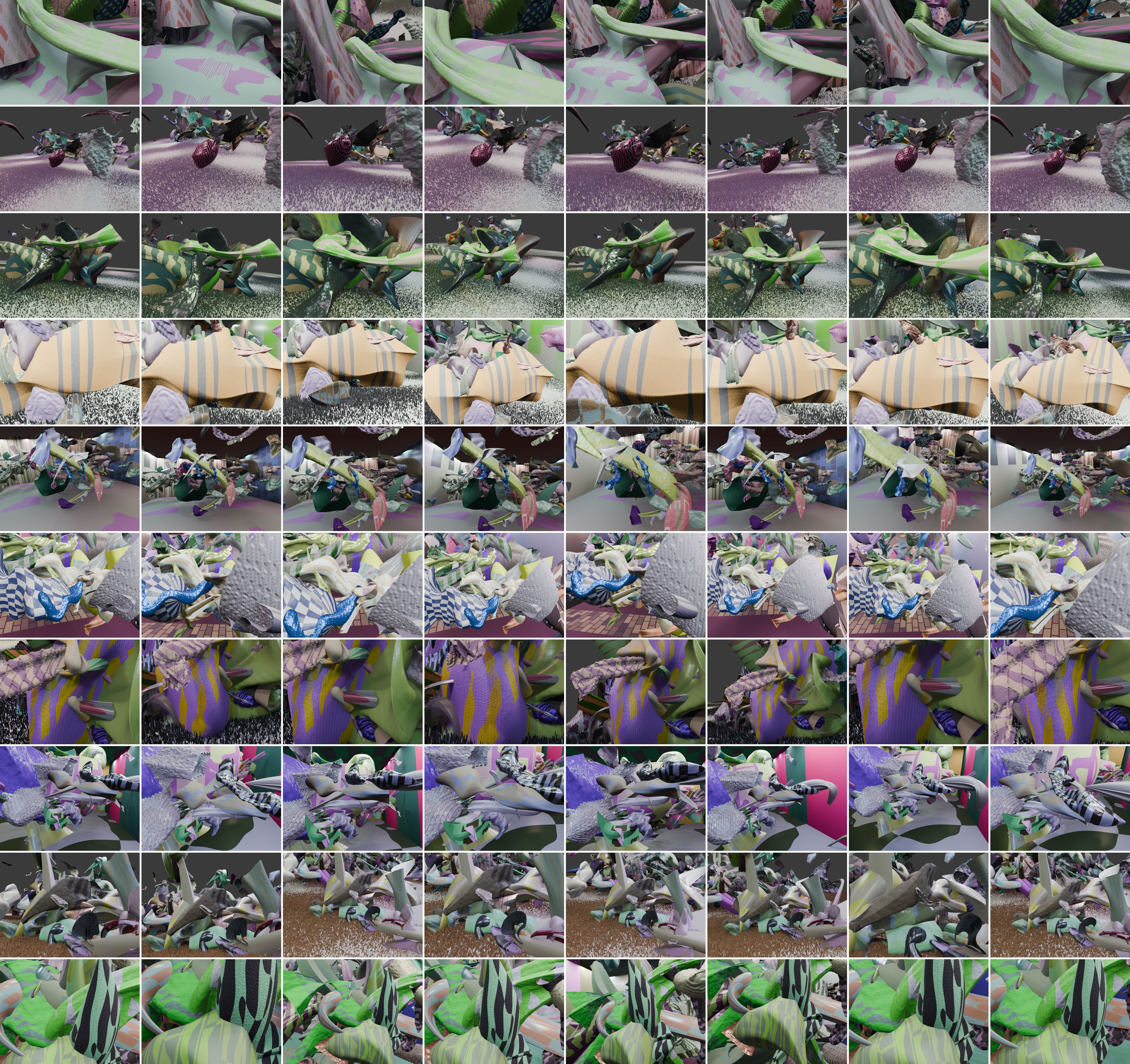}
  \caption{Gallery of 10 random examples of our generated scenes}
  \label{fig:scenes}
\end{figure}

\subsection{Object Arrangement, Lighting and Rendering}

We place $n_0$ large objects and $n_1$ small objects in the scene. One of the large objects is always in the center as the main object. For the other large objects, we ensure they are visible in at least half of the cameras.

After the large objects are placed, we place small objects in a mixture of two ways:
\begin{itemize}
    \item Uniform: We compute the bounding box of all the large objects and place a small object uniformly within it.
    \item Clustered: We sample a large object and sample a surface point on it to place a small object. This mimics realistic placement and contact between objects.
\end{itemize}

Optionally, we add an enclosing room box with a 50\% probability to simulate an indoor environment with large planar surfaces. We apply textures (the same as Sec.~\ref{sec:tex}) to the room box, but not displacement. We also optionally scatter small objects on a ground plane, again with a 50\% probability, to mimic an outdoor grass field.

Area lights are placed on a plane above the scene. When a room box is present, lights are placed within the room. For rendering efficiency, we render using the EEVEE engine in Blender.

Fig.~\ref{fig:scenes} shows 10 random examples of our generated scenes. Each row contains 8 cameras in the same scene with suitable covisibility.  We can see ground-scattered objects in rows 2, 3, etc, and room boxes in rows 4, 5, etc.

Please see the \textbf{ablation experiments} in the Appendix for a justification of the chosen design decisions.

\section{Experiments}

\paragraph{\textbf{Scene Generation}}
We render our dataset on a cluster.
Scene generation takes 490~s per scene on average on a single CPU core with 64 GB RAM, and rendering takes 262~s per image on an NVIDIA RTX 3090. Generation is fully parallel across scenes and the render jobs use various NVIDIA GPU models (including RTX 2080Ti and RTX 3090).

\paragraph{\textbf{Training and Evaluation}}

We train the MVSAnywhere model~\cite{izquierdo2025mvsanywhere} and evaluate the model following RMVD benchmark~\cite{schroppel2022benchmark}. We split each dataset equally into \textbf{a validation set} and \textbf{a test set}. We optimize our data generator using the validation set and we report the score on the test set. For ScanNet, we use the frame selection from the MVSAnywhere paper. We also undistort both the images and the depth maps in ETH3D dataset as pointed out in that paper. The training and evaluation were performed using NVIDIA RTX 3090 and RTX A6000 GPUs.

We run each training process \textbf{3 times} and report the mean performance for the relative error ($\text{rel}$) and threshold accuracy ($\tau$) across each benchmark and the aggregate score; we also report the  \textbf{standard deviation} of the performance on the aggregate score.

\subsection{Fixed-Budget Comparison}

\label{sec:same_budget}
To isolate data quality from dataset size, we fix the budget to be 8,000 images. We maximize the total number of scenes as this provides the most diversity. A minimum of 8 images per scene is required for MVSAnywhere to train in a multi-view setting. We select such a subset of 1000 scenes and 8 images per scene from (a) training set of MVSAnywhere (8-dataset) and (b) MegaSynth~\cite{jiang2025megasynth}. We generate an equivalent amount of 8,000 images with SimpleProc, which takes approximately 580 GPU-hours.

When we sample images from the 8-datasets mixture, we preserve the original dataset proportions to ensure the best performance under the current budget. MegaSynth does not contain covisibility information for pair selection in multiview stereo. Therefore, we select the images based on the following rule: given one camera view and an inter-camera angle threshold, we identify other views within this threshold and select the closest 7 views as its neighbor views. We randomly pick 1000 such tuples in 1000 scenes. We optimize the threshold on the validation split and report the results with the best threshold on the test split. 

We performed three training runs of 200,000 steps each with a batch size of 3 and show the results in Table~\ref{tab:same_budget}.

\begin{table}[h!]
\centering
\small
\setlength{\tabcolsep}{4pt}
\resizebox{\textwidth}{!}{
\begin{tabular}{l|cc|cc|cc|cc|cc|cc}
\toprule
\textbf{Test Split} 
& \multicolumn{2}{c|}{KITTI} 
& \multicolumn{2}{c|}{ScanNet} 
& \multicolumn{2}{c|}{ETH3D} 
& \multicolumn{2}{c|}{DTU} 
& \multicolumn{2}{c|}{T\&T} 
& \multicolumn{2}{c}{Average} \\
& rel $\downarrow$ & $\tau \uparrow$
& rel $\downarrow$ & $\tau \uparrow$
& rel $\downarrow$ & $\tau \uparrow$
& rel $\downarrow$ & $\tau \uparrow$
& rel $\downarrow$ & $\tau \uparrow$
& rel $\downarrow$ & $\tau \uparrow$ \\
\midrule
8-dataset & 4.56 & 55.68 & \textbf{4.38} & \textbf{56.83} & 4.32 & 80.54 & 1.37 & 93.39 & 2.96 & 79.54 & 3.52 ($\pm$ 0.03) & 73.20 ($\pm$ 0.11) \\

MegaSynth & 4.69 & 54.09& 5.97& 49.62& 4.71 & 79.70& 3.05& 94.00& 3.65& 78.26& 4.41 ($\pm$ 0.42) & 71.13 ($\pm$ 0.24)\\
Ours & \textbf{3.41}& \textbf{65.64}& 5.19& 53.29& \textbf{3.98}& \textbf{85.57}& \textbf{0.84}&\textbf{ 96.05}& \textbf{2.93}& \textbf{85.12}& \textbf{3.27} ($\pm$ 0.06)& \textbf{77.13 }($\pm$ 0.20)\\
\bottomrule
\end{tabular}
}

\vspace{2mm}
\caption{Performance of models trained on a fixed budget of 8,000 images. Our procedural data demonstrates superior performance across nearly all benchmarks, with the exception of ScanNet.}
\label{tab:same_budget}
\end{table}

Our model achieves significantly better results on all the benchmarks except for ScanNet, resulting in a much higher average score, showing the data efficiency of our approach.

Note that ScanNet is different from other datasets in two aspects: first, it contains noisier ground truth due to sensor limitations; second, many keyframe regions lack covisibility in neighboring views, which forces the model to rely on monocular priors rather than geometric 3D correspondences, favoring models trained on the 8-dataset mix due to its similar indoor priors.

\begin{figure}[h!]
\centering
\includegraphics[width=1\linewidth]{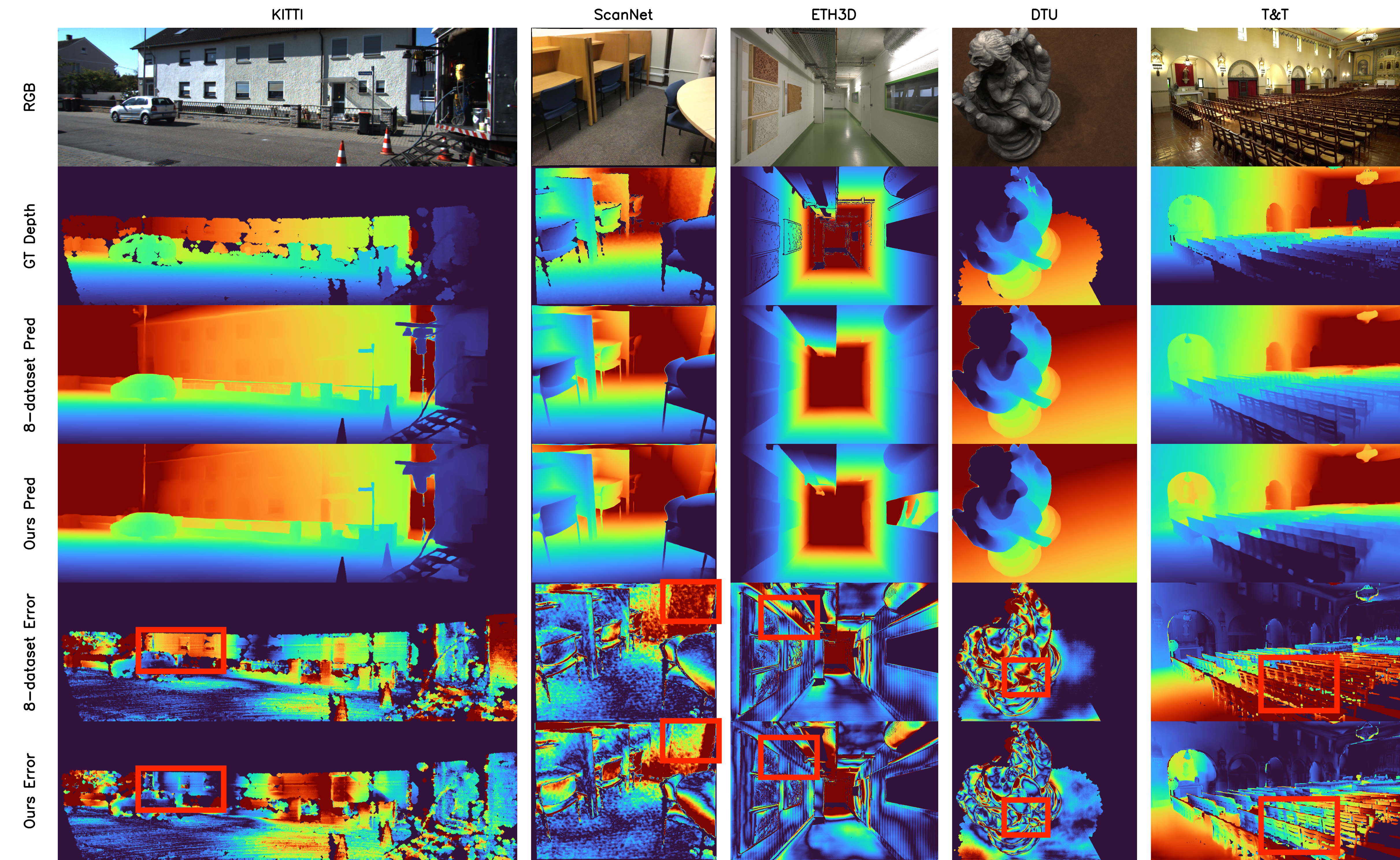}

\includegraphics[width=1\linewidth]{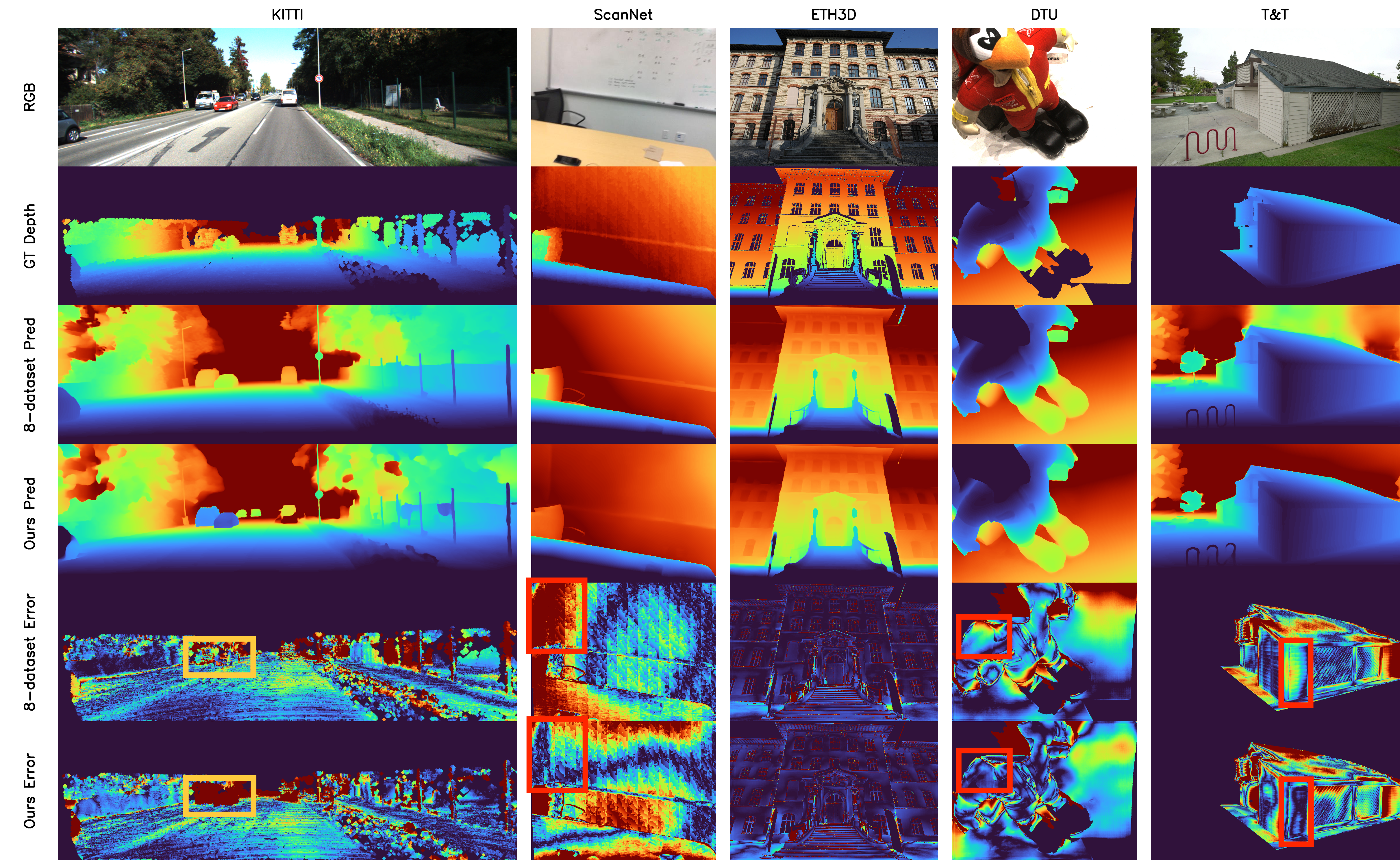}
\caption{Unlimited-Budget qualitative comparison across five benchmarks. Our procedural training data yields competitive depth estimation results compared to the 8-dataset baseline.}
\label{fig:qualitative_results}
\end{figure}

\subsection{Unlimited-Budget Comparison}

Next, we compare performance of our data against the original MVSAnywhere training data. We rendered 352,000 images in total using our procedural generator, which takes about 25,600 GPU-hours, including variant configurations during generator development. We performed three training runs of 1,600,000 steps each with a batch size of 6 and show the results in Table~\ref{tab:unlimited}. In addition to the standard 5-benchmark average, we compute an average excluding ScanNet to evaluate performance on more reliable and typical benchmarks (for the reasons discussed in Sec.~\ref{sec:same_budget}). We report results for both the test split and the entire dataset for reference, though the test split is the primary rigorous metric. 

We excluded MegaSynth from this experiment as it has very high storage cost for full-scale training, and showed lower performance in the previous experiment. For the 8-dataset configuration, we use the off-the-shelf model (trained on over 692,000 images) rather than retraining (no standard deviation is reported for this data point).

\begin{table}[h!]
\centering
\small
\setlength{\tabcolsep}{3pt}
\newcommand{\std}[1]{($\pm$#1)} %

\resizebox{\textwidth}{!}{
\begin{tabular}{l|cc|cc|cc|cc|cc|cc|cc}
\toprule
\textbf{Test Split} 
& \multicolumn{2}{c|}{KITTI} 
& \multicolumn{2}{c|}{ScanNet} 
& \multicolumn{2}{c|}{ETH3D} 
& \multicolumn{2}{c|}{DTU} 
& \multicolumn{2}{c|}{T\&T} 
& \multicolumn{2}{c|}{\textbf{Average}}
& \multicolumn{2}{c}{\textbf{Avg ex-S}} \\
& rel $\downarrow$ & $\tau \uparrow$
& rel $\downarrow$ & $\tau \uparrow$
& rel $\downarrow$ & $\tau \uparrow$
& rel $\downarrow$ & $\tau \uparrow$
& rel $\downarrow$ & $\tau \uparrow$
& rel $\downarrow$ & $\tau \uparrow$ 
& rel $\downarrow$ & $\tau \uparrow$ \\
\midrule
8-dataset & \textbf{2.98} & 71.02 & \textbf{3.70} & \textbf{64.04} & 3.52 & 88.58 & 0.86 & \textbf{96.90} & \textbf{2.21} & \textbf{88.49} & \textbf{2.66} & \textbf{81.80} & 2.39 & 86.25 \\
Ours & 3.00 & \textbf{71.60} & 5.06 & 53.47 & \textbf{3.35} & \textbf{90.36} & \textbf{0.75} & 96.58 & 2.43 & 87.92 & 2.92 & 79.99 & \textbf{2.38} & \textbf{86.62} \\
& & & & & & & & & & & \std{0.02} & \std{0.06} & \std{0.03} & \std{0.20} \\
\bottomrule
\end{tabular}
}

\vspace{1em} %

\resizebox{\textwidth}{!}{
\begin{tabular}{l|cc|cc|cc|cc|cc|cc|cc}
\toprule
\textbf{Both splits} 
& \multicolumn{2}{c|}{KITTI} 
& \multicolumn{2}{c|}{ScanNet} 
& \multicolumn{2}{c|}{ETH3D} 
& \multicolumn{2}{c|}{DTU} 
& \multicolumn{2}{c|}{T\&T} 
& \multicolumn{2}{c|}{\textbf{Average}}
& \multicolumn{2}{c}{\textbf{Avg ex-S}} \\
& rel $\downarrow$ & $\tau \uparrow$
& rel $\downarrow$ & $\tau \uparrow$
& rel $\downarrow$ & $\tau \uparrow$
& rel $\downarrow$ & $\tau \uparrow$
& rel $\downarrow$ & $\tau \uparrow$
& rel $\downarrow$ & $\tau \uparrow$
& rel $\downarrow$ & $\tau \uparrow$ \\
\midrule
8-dataset & \textbf{3.24} & \textbf{68.80} & \textbf{3.69} & \textbf{65.09} & \textbf{3.32} & \textbf{88.75} & 1.27 & \textbf{94.98} & \textbf{2.13} & \textbf{90.48} & \textbf{2.73} & \textbf{81.62} & \textbf{2.49} & \textbf{85.75} \\
Ours & 3.39 & 67.84 & 4.95 & 55.49 & 3.43 & 88.67 & \textbf{1.09} & 94.20 & 2.42 & 89.86 & 3.06 & 79.21 & 2.58 & 85.14 \\
& & & & & & & & & & & \std{0.03} & \std{0.17} & \std{0.04} & \std{0.20} \\
\bottomrule
\end{tabular}
}
\vspace{2mm}
\caption{Comparing the 8-dataset baseline with Ours under an unlimited budget. Excluding ScanNet (Avg ex-S), our method achieves superior average performance on the Test Split.}
\label{tab:unlimited}
\end{table}

Interestingly, our advantage is more pronounced on the test set, confirming that the generator does not overfit the validation split.
The test set results show that we achieved comparable or sometimes even better scores across all benchmarks except ScanNet. Notably, this approach outperforms the "8-dataset" baseline on the KITTI $\tau$ metric, both ETH3D metrics, and the DTU $\text{rel}$ metric. Excluding ScanNet, our model achieves a 0.4\% improvement on both $\tau$ and $\text{rel}$. Including ScanNet, our model is only 2.2\% worse on $\tau$.

In Figure \ref{fig:qualitative_results}, we show two random samples in each of the 5 benchmarks. We show the RGB image of the reference view (7 neighbor views are omitted), the ground-truth depth, prediction of the model trained from 8-datasets and prediction from the model trained on our data. Each example shows competitive results, with clear regions where our model performs better (in red boxes).
The circle/stripe artifacts in the error map of ScanNet likely suggest RGB-depth desynchronization and inaccurate calibration in the ground-truth. In KITTI (yellow boxes), moving cars violate the static multi-view assumption. Our model treats the car as static and was misled. This may mean that changing only the training data of MVSAnywhere results in a greater focus on 3D correspondences rather than memorizing domain-specific priors, improving generalization ability.

Both of these quantitative and qualitative results demonstrate that procedural training data with a small set of rules is effective for multi-view stereo.

\subsection{Limitation and Future Work}

The results, especially on ScanNet, could potentially be improved if we incorporate a small amount of real-world indoor data as priors. Furthermore, SimpleProc currently does not model dynamic objects, sensor noise, lens effects, or complex materials such as strongly specular and transparent surfaces. We leave these as future work.

\section{Conclusion}
In this paper, we built SimpleProc, a procedural data generation system based on NURBS and simple texture patterns and demonstrated that we can generate effective training data for multi-view stereo with fully procedural data from a small set of optimized rules.

\section*{Acknowledgements}
This work was partially supported by the National Science Foundation.
{
\small

\bibliographystyle{abbrv} %
\bibliography{main} %
}

\newpage
\appendix

\section{Ablation Study}

\label{sec:abl}
We ablate the design of object shapes, room boxes, materials, displacements, object sizes, lighting, and camera settings. The results are presented in Tables~\ref{tab:abl_part1} and~\ref{tab:abl_part2}.

\paragraph{\textbf{Experimental Setup}}
In each experimental block, only one feature is modified while others are kept constant within that block (including the random seed). While these fixed parameters may differ between blocks, we assume that the observed trends are independent of these variations. We observed that the advantage might shift between configurations as the dataset size grows. Therefore, we chose 1,000 scenes in total to ensure relatively stable results.

We perform three training runs of 200,000 steps each with a batch size of 3. We report the mean performance across all benchmarks and for the aggregate score, including the standard deviation of the aggregate score. We focus exclusively on the $\tau$ metric, as it is more robust.

\begin{table}[b!]
\centering
\small
\setlength{\tabcolsep}{4pt}
\resizebox{0.8\textwidth}{!}{
\begin{tabular}{l|c|c|c|c|c|c}
\toprule
\textbf{Validation Split} 
& KITTI 
& ScanNet 
& ETH3D 
& DTU 
& T\&T 
& Average \\
    & $\tau \uparrow$ & $\tau \uparrow$ & $\tau \uparrow$ & $\tau \uparrow$ & $\tau \uparrow$ & $\tau \uparrow$ \\
\midrule

\multicolumn{7}{l}{\textbf{Shape Type}} \\
 Primitives & 55.79 & 51.56 & 77.24 & 84.22 & 90.25 & 71.81 ($\pm$ 0.86) \\
 Primitives + NURBS & 60.12 & 53.54 & 81.55 & 88.37 & 92.32 & 75.18 ($\pm$ 0.54) \\
 NURBS & \textbf{61.53} & \textbf{55.76} & \textbf{83.64} & \textbf{89.61} & \textbf{93.28} & \textbf{76.77} ($\pm$ 0.21) \\

\midrule

\multicolumn{7}{l}{\textbf{Displacement}} \\
 No & 61.37 & \textbf{56.17} & 80.24 & 86.32 & 91.81 & 75.18 ($\pm$ 0.16) \\
 Yes  & \textbf{62.62} & 56.15 & \textbf{81.21} & \textbf{89.79} & \textbf{92.20} & \textbf{76.40} ($\pm$ 0.50) \\

\midrule

\multicolumn{7}{l}{\textbf{Materials}} \\
 Uniform Color              & 60.03 & 55.62 & 76.54 & 85.89 & 91.13 & 73.84 ($\pm$ 0.46) \\
 w/ Noise Texture           & \textbf{62.29} & \textbf{55.90} & 78.91 & 87.85 & \textbf{92.57} & 75.51 ($\pm$ 0.04) \\
 w/ Noise Texture + Boolean & 62.07 & 55.87 & \textbf{80.21} & \textbf{88.43} & 92.34 & \textbf{75.78} ($\pm$ 0.37) \\

\midrule

\multicolumn{7}{l}{\textbf{Number of Large Objects}} \\
 1 & 57.15 & 52.11 & 70.08 & 83.29 & 90.55 & 70.64 ($\pm$ 1.20) \\
 2 & 59.01 & 52.51 & 76.22 & 83.68 & 92.05 & 72.69 ($\pm$ 0.67) \\
 8 & \textbf{62.72} & \textbf{55.84} & \textbf{79.82} & \textbf{87.99} & \textbf{93.05} & \textbf{75.88} ($\pm$ 0.23) \\

\midrule

\multicolumn{7}{l}{\textbf{Small Objects}} \\
 None               & 60.71 & 55.82 & 81.43 & 86.49 & 91.68 & 75.23 ($\pm$ 0.57) \\
 Uniform Placement       & 61.12 & \textbf{56.63} & 81.22 & \textbf{90.23} & 91.46 & 76.13 ($\pm$ 0.16) \\
 Clustered Placement    & \textbf{62.08} & 55.42 & 80.33 & 88.84 & \textbf{92.51} & 75.84 ($\pm$ 0.13) \\
 50-50 Clustered and Uniform & 61.73 & 55.68 & \textbf{81.97} & 89.71 & 92.06 & \textbf{76.23} ($\pm$ 0.28) \\

\midrule

\multicolumn{7}{l}{\textbf{Small Objects: Number and Size}} \\
 160;  Smaller & 60.86 & 56.40 & 81.92 & 90.21 & 91.59 & 76.20 ($\pm$ 0.32) \\
 160;  Larger  & \textbf{61.49} & 56.23 & \textbf{83.23} & 90.03 & 91.98 & 76.59 ($\pm$ 0.42) \\
 320;  Smaller & 61.12 & 56.63 & 81.22 & 90.23 & 91.46 & 76.13 ($\pm$ 0.16) \\
 320;  Larger  & 61.47 & \textbf{57.05} & 83.19 & \textbf{90.94} & \textbf{92.25} & \textbf{76.98} ($\pm$ 0.13) \\
\midrule

\multicolumn{7}{l}{\textbf{Room Box}} \\
 w/o & 60.52 & 53.03 & 75.52 & \textbf{88.94} & 92.47 & 74.10 ($\pm$ 0.40) \\
 w/  & \textbf{62.12} & \textbf{55.47} & \textbf{78.40} & 88.83 & \textbf{92.59} & \textbf{75.48} ($\pm$ 0.32) \\

\midrule

\multicolumn{7}{l}{\textbf{Scattered Tiny Objects}} \\
 w/o & 61.20 & 54.24 & \textbf{80.68} & 86.60 & 92.31 & 75.01 ($\pm$ 0.38) \\
 w/  & \textbf{62.12} & \textbf{55.47} & 78.40 & \textbf{88.83} & \textbf{92.59} & \textbf{75.48} ($\pm$ 0.32) \\
 
\bottomrule
\end{tabular}
}

\vspace{2mm}
\caption{Ablation study on procedural rule design choices for object geometry, surface materials, and scene arrangement.}
\label{tab:abl_part1}
\end{table}

\paragraph{\textbf{Selection Criteria}}
While individual ablation runs exhibit some noise, we follow the principle that multiple incremental improvements can accumulate in the final result. Therefore, in general, we select the setting with the best aggregate score regardless of the standard deviation. 
For some cost-sensitive parameters, e.g., the number of large objects, we cap parameter values to maintain data generation efficiency.

\paragraph{\textbf{Individual Analysis}}
\begin{itemize}
    \item \textbf{Shape Type}: Instead of our proposed lofting-based NURBS shapes, we evaluated basic primitives, i.e., a mixture of cubes, spheres, and cylinders. We also tried a mixture of NURBS shapes and primitives. The proposed NURBS shapes achieve the best accuracy.

    \item \textbf{Displacements}: We compared the configuration with and without our noise-based displacement. The results show that the proposed displacement constantly improves the performance.

    \item \textbf{Materials}: We compared three configurations: (a) uniform (single color) for each object, (b) with noise texture but without the proposed Boolean modification, and (c) with noise texture and the boolean modification. Configuration (c) achieves the best score.

    \item \textbf{Number of Objects}: We evaluated the effect of the number of large objects (scaling small objects proportionally). The results show that increasing object count improves the performance, but we cap at 8 large objects to maintain practical efficiency.

    \item \textbf{Small Objects}: We compared several configurations: None, Uniform, Clustered (around large objects), and Hybrid (Uniform + Clustered). The results show that the Hybrid configuration performs best. We also evaluated the impact of number and size of the small objects. The results show that increasing both the count and scale of small objects improves the performance.
    
    \item \textbf{Room Box}: We compared the configurations with and without the room box. The room box improves the performance on all the benchmarks, especially for ETH3D.

    \item \textbf{Scattered Tiny Objects}: We compared the configurations with and without the scattered tiny objects. These objects improve the performance on all the benchmarks except ETH3D, likely because the ground planes in ETH3D are mostly flat.

    \item \textbf{Lighting}: We evaluated the impact of the number of lights, including random configurations. The results show that increasing the number of lights improves the performance, but we cap at 80 lights in total.
    
    \item \textbf{Cameras}: Our default camera setting uses randomized ranges for FoV, camera-to-center distance, and inter-camera azimuth. We compared this against fixed settings for FoV and distance, as well as smaller or larger azimuth ranges. The results show that our default random range performs best.

\end{itemize}

\begin{table}[b!]
\centering
\small
\setlength{\tabcolsep}{4pt}
\resizebox{0.8\textwidth}{!}{
\begin{tabular}{l|c|c|c|c|c|c}
\toprule
\textbf{Validation Split} 
& KITTI 
& ScanNet 
& ETH3D 
& DTU 
& T\&T 
& Average \\
    & $\tau \uparrow$ & $\tau \uparrow$ & $\tau \uparrow$ & $\tau \uparrow$ & $\tau \uparrow$ & $\tau \uparrow$ \\
\midrule

\multicolumn{7}{l}{\textbf{Number of Lights}} \\
 5 - 10 & 62.17 & 55.57 & 81.52 & 87.83 & 92.52 & 75.92 ($\pm$ 0.19) \\
 5 - 20 & 62.44 & 55.96 & \textbf{81.70} & 87.83 & 92.10 & 76.01 ($\pm$ 0.17) \\
 5 - 40 & \textbf{62.49} & 55.23 & 81.37 & 86.65 & 92.27 & 75.60 ($\pm$ 0.51) \\
 5 - 80 & 62.48 & 55.86 & 80.92 & 89.34 & 92.71 & 76.26 ($\pm$ 0.10) \\
 40     & \textbf{62.49} & 56.07 & 80.14 & 88.25 & 92.71 & 75.93 ($\pm$ 0.16) \\
 80     & 62.13 & \textbf{57.06} & 80.21 & \textbf{90.59} & \textbf{93.36} & \textbf{76.67} ($\pm$ 0.35) \\
 
\midrule

\multicolumn{7}{l}{\textbf{Camera Settings}} \\
 Constant Small FoV  & 57.04 & 50.52 & 66.22 & 83.33 & 89.75 & 69.37 ($\pm$ 0.30) \\
 Constant Medium FoV & 60.44 & 53.19 & 74.96 & 88.99 & 92.91 & 74.10 ($\pm$ 0.12) \\
 Constant Large FoV  & 57.86 & 51.24 & 74.32 & 83.28 & 92.13 & 71.77 ($\pm$ 0.49) \\
 w/o Distance Change & 61.66 & 53.03 & 76.50 & 87.50 & 92.68 & 74.27 ($\pm$ 0.33) \\
 Larger Azimuth Range  & 59.60 & \textbf{56.26} & 78.70 & 80.07 & 90.89 & 73.10 ($\pm$ 0.79) \\
 Smaller Azimuth Range & 62.45 & 54.93 & 79.62 & \textbf{89.16} & 92.09 & 75.65 ($\pm$ 0.15) \\
 Default              & \textbf{62.72} & 55.84 & \textbf{79.82} & 87.99 & \textbf{93.05} & \textbf{75.88} ($\pm$ 0.23) \\
 
\bottomrule
\end{tabular}
}
\vspace{2mm}
\caption{Ablation study on lighting and camera parameters.}

\label{tab:abl_part2}
\end{table}

\end{document}